\documentclass{article}

\usepackage[preprint]{neurips_2026}

\usepackage{enumitem}
\usepackage[hidelinks]{hyperref}
\usepackage{url}
\usepackage{booktabs}
\usepackage{amsfonts}
\usepackage{graphicx}
\usepackage{fontawesome5}
\usepackage[small]{caption}
\usepackage{subcaption}
\usepackage{amsmath}
\allowdisplaybreaks
\usepackage{wrapfig}
\usepackage{booktabs}
\usepackage{multicol}

\usepackage{float}
\usepackage{xcolor}
\usepackage{colortbl}
\usepackage{amssymb}
\usepackage{adjustbox}
\usepackage{pifont}
\usepackage[noabbrev,capitalize]{cleveref}

\usepackage[compact]{titlesec}
\titlespacing{\section}{0pt}{1ex}{0.5ex}
\titlespacing{\subsection}{0pt}{0.5ex}{0ex}
\titlespacing{\subsubsection}{0pt}{0.5ex}{0ex}
\usepackage{setspace}
\AtBeginDocument{%
  \addtolength\abovedisplayskip{-0.25\baselineskip}%
  \addtolength\belowdisplayskip{-0.25\baselineskip}%
  \addtolength\abovedisplayshortskip{-0.25\baselineskip}%
  \addtolength\belowdisplayshortskip{-0.25\baselineskip}%
}

\usepackage{xspace}
\newcommand{\circone}{\ding{172}\xspace}
\newcommand{\circtwo}{\ding{173}\xspace}
\newcommand{\circthree}{\ding{174}\xspace}

\usepackage{tabularx, booktabs}
\newcolumntype{C}{>{\centering\arraybackslash}X}
\newcolumntype{R}{>{\raggedleft\arraybackslash}X}
\newcolumntype{S}{>{\raggedleft\arraybackslash\hsize=.5\hsize}X}

\usepackage{multirow}
\usepackage{makecell}

\crefname{equation}{equation}{equations}
\crefname{section}{section}{sections}
\crefname{figure}{Figure}{Figures}
\crefname{table}{Table}{Tables}
\crefname{appendix}{Appendix}{Appendices}
\Crefname{appendix}{Appendix}{Appendices}
\crefname{lstlisting}{Listing}{Listings}
\Crefname{lstlisting}{Listing}{Listings}
\usepackage{appendix}
\usepackage{listings}
\newcommand{\codefont}{\fontfamily{lmtt}\selectfont}
\lstdefinestyle{datastyle}{
  basicstyle={\codefont\small},
  xleftmargin={6pt},
  xrightmargin={6pt},
  breakindent=0pt,
  frame=tb,
  tabsize=2,
  showspaces=false,
  showstringspaces=false,
  extendedchars=true,
  breaklines=true,
  columns=fullflexible,
  keepspaces=true,
  captionpos=b,
  breakatwhitespace=false,
  numbersep=5pt,
  aboveskip={0.8\baselineskip},
  belowskip={0.2\baselineskip},
  backgroundcolor=\color{aigreen},
  escapeinside={@}{@},
}
\usepackage{tikz}
\usetikzlibrary{arrows.meta}
\usepackage{pgfplots}
\pgfplotsset{compat=1.18}
\usepgfplotslibrary{groupplots}

\definecolor{aigreen}{RGB}{245, 255, 249}

\usepackage[textwidth=2.7cm]{todonotes}

\newcommand\myshade{85}
\colorlet{myurlcolor}{blue}
\hypersetup{
  citecolor  = black,
  urlcolor   = myurlcolor!\myshade!black,
  colorlinks = true,
}

\newcommand{\modelname}{\textsc{ZooClaw-FashionSigLIP2}\xspace}
\newcommand{\wiseft}{\textsc{WiSE-FT}\xspace}
\newcommand{\modelsoup}{model soup\xspace}

\title{ZooClaw-FashionSigLIP2: Distilled Fine-tuning for Robust Fashion Retrieval}

\author{%
  Siqiao Xue, Chunxue Xu \\ [0.6em]
  ZooClaw.ai \\[0.6em]
  \textbf{\href{https://huggingface.co/srpone/zooclaw-fashionsiglip2}
     {\textcolor{blue!60!black}{\faIcon{robot}\enspace{Model}}}
    \quad
  \href{https://huggingface.co/datasets/srpone/zooclaw-fashion-eval}
     {\textcolor{blue!60!black}{\faDatabase\enspace{Dataset}}}
     \quad
    \href{https://github.com/SerendipityOneInc/look-bench}%
      {\textcolor{blue!60!black}{\faGithub\enspace Code}}%
  }%
}

\begin{document}

\maketitle

\begin{abstract}
Adapting a foundation vision-language encoder to a specialized retrieval task creates a fundamental tradeoff: gains on the target distribution come at the cost of the foundation model's broad generalization, and fashion retrieval is a stringent instance of this problem.
We present \modelname, a fashion-specialized SigLIP2-base model that resolves this tradeoff with a simple recipe --- full fine-tuning with knowledge distillation on curated in-domain data, followed by \wiseft~\citep{wortsman2022wiseft} weight interpolation with the base model --- and outperforms LoRA, larger backbones (up to 1B parameters), and external training data.
Under fair evaluation, \modelname outperforms all baselines on every benchmark in our suite.
In addition, we release ZooClaw-Fashion, a new high-quality fashion retrieval benchmark, and a systematic quality analysis of widely-used benchmarks that exposes and mitigates structural biases in their public ground truth.
We open-source the model weights and all evaluation artifacts to facilitate future research.
\end{abstract}

\section{Introduction}
\label{sec:intro}

Fashion image-text retrieval, the task of matching product images to natural language queries and vice versa, is a core component of modern e-commerce search and recommendation systems.
In production, the vast majority of search traffic consists of short keyword queries (e.g., ``red floral maxi dress''), yet most vision-language encoder (VLE) benchmarks and training recipes focus on detailed natural-language descriptions.
A practical fashion retrieval model must excel at both: short queries that users actually type and richer descriptions used in catalog systems.
Pre-trained VLEs like CLIP~\citep{radford2021clip}, SigLIP~\citep{zhai2023siglip}, and SigLIP2~\citep{tschannen2025siglip2} provide strong zero-shot embeddings, but their general-purpose training does not capture the fine-grained visual and textual distinctions that fashion retrieval demands~\citep{gao2026lookbench}: subtle differences in neckline, fabric texture, or styling details that determine relevance.

Domain-specific fine-tuning is the natural remedy, and models like Marqo-fashionCLIP demonstrate large in-domain gains through contrastive training on fashion data.
However, fine-tuning introduces a fundamental tension between in-domain performance and out-of-distribution (OOD) generalization: a production system must serve not only its training catalog but also unseen public catalogs with different query styles, such as structured attribute queries in Fashion200k~\citep{han2017fashion200k} or catalog product descriptions in H\&M~\citep{hm2022kaggle}.

In this work, we present \modelname, a fashion-specialized VLE that resolves this tension through distilled fine-tuning.
Through systematic experimentation on the SigLIP2 family, we identify an effective recipe: full fine-tuning with knowledge distillation on curated in-domain data, followed by \wiseft~\citep{wortsman2022wiseft} weight interpolation between the fine-tuned and base checkpoints.
Neither parameter-efficient methods (LoRA), nor scaling to larger backbones (up to 1B parameters), nor augmenting with external data matches this recipe.

Under fair evaluation, \modelname leads or ties on every metric of every benchmark in our suite, outperforming Marqo-fashionCLIP, Marqo-fashionSigLIP, and the zero-shot SigLIP2 family.
For Fashion200k, we adopt a TREC-style pooled re-evaluation with 102{,}494 held-out judgments, after finding that its public ground truth is biased toward caption-source instance recovery rather than relevance.
We open-source the model weights, the ZooClaw-Fashion benchmark, and our pooled evaluation artifacts to facilitate future research.
A continuously optimized commercial version is available at \url{https://zoodata.ai/en/api-docs}.

\section{Related Work}
\label{sec:related}

\paragraph{VLEs for retrieval.}
CLIP~\citep{radford2021clip} established the VLE paradigm for image-text retrieval via contrastive pre-training on web-scale data.
Subsequent work improved architecture (SigLIP~\citep{zhai2023siglip}, SigLIP2~\citep{tschannen2025siglip2}), training efficiency (OpenCLIP~\citep{ilharco2021openclip}), and domain specialization (FashionCLIP~\citep{chia2024fashionclip}).
SigLIP2 replaces the softmax-based InfoNCE loss with a sigmoid loss that eliminates the need for large batch sizes and supports multi-task training.

\paragraph{Fashion retrieval.}
Fashion-specific models include FashionCLIP~\citep{chia2024fashionclip}, Marqo-fashionCLIP~\citep{marqo2024fashionclip}, and Marqo-fashionSigLIP~\citep{marqo2024fashionsiglip}.
These models improve in-domain performance but often sacrifice generalization.
Benchmarks include DeepFashion~\citep{liu2016deepfashion}, Fashion200k~\citep{han2017fashion200k} and LookBench~\citep{gao2026lookbench}.


\paragraph{Model soups and weight-space ensembling.}
Model soups~\citep{wortsman2022modelsoup} average the weights of multiple fine-tuned models to improve accuracy without increasing inference cost.
\wiseft~\citep{wortsman2022wiseft} is a special case that interpolates between the zero-shot and a single fine-tuned checkpoint to improve OOD robustness.
These methods exploit the observation that fine-tuned models often lie in the same loss basin as the pre-trained model, enabling linear combinations that retain in-domain gains while recovering OOD performance.

\section{Method}
\label{sec:method}

\subsection{Problem Formulation}
\label{sec:problem}

Given a corpus of product images $\mathcal{V} = \{v_1, \ldots, v_M\}$ and a set of text queries $\mathcal{T} = \{t_1, \ldots, t_N\}$, fashion image-text retrieval aims to learn an embedding space where relevant image-text pairs have high cosine similarity.
We consider two retrieval directions: text-to-image (T2I), where a text query retrieves the most relevant images, and image-to-text (I2T), the reverse.
A key challenge in practice is that queries vary widely in form---from short keywords (e.g., ``red floral maxi dress'') to detailed natural-language descriptions---and the model must generalize across both in-domain and out-of-distribution evaluation scenarios.

We adopt a VLE architecture consisting of an image encoder $f_\theta$ and a text encoder $g_\phi$ that map inputs to a shared $d$-dimensional embedding space.
Our goal is to fine-tune a pre-trained VLE (SigLIP2-base) for fashion retrieval while preserving its OOD generalization.
Our approach consists of three stages: \circone multi-task contrastive fine-tuning with knowledge distillation (\cref{sec:training}), \circtwo \wiseft weight interpolation (\cref{sec:wiseft}), and \circthree selection of the interpolation coefficient.

\subsection{Multi-Task Contrastive Training}
\label{sec:training}

\paragraph{Full fine-tuning over LoRA.}
We use full fine-tuning rather than parameter-efficient methods such as LoRA.
Full-rank updates provide the capacity needed to absorb the multi-task objective and the distillation regularizer (introduced below), and our ablations confirm that no LoRA configuration matches full fine-tuning on our suite (see \cref{sec:main_results}).

\paragraph{Training objective.}
We train with a Generalized Contrastive Loss (GCL)~\citep{zhang2022contrastive} that extends InfoNCE~\citep{oord2018infonce} by incorporating graded relevance scores, allowing the model to leverage soft labels rather than treating all non-matching pairs as equally negative.
Given a batch of $N$ image-text pairs $\{(v_i, t_i, r_i)\}$ where $r_i \in [0,1]$ is the relevance score of pair $i$, the text-to-image loss is:
\begin{equation}
\mathcal{L}_{\text{t2i}} = -\frac{1}{N} \sum_{i=1}^{N} \log \frac{\exp(\text{sim}(t_i, v_i) / \tau)}{\sum_{j=1}^{N} w_{ij} \cdot \exp(\text{sim}(t_i, v_j) / \tau)},
\label{eq:gcl}
\end{equation}
where $\text{sim}(\cdot,\cdot)$ denotes cosine similarity, $\tau$ is a learnable temperature, and $w_{ij} = 1 - r_j \cdot \mathbb{1}[i \neq j]$ downweights in-batch negatives drawn from pairs that are themselves highly relevant matches --- such items are likely false negatives and should not be pushed apart with full strength.
The image-to-text counterpart $\mathcal{L}_{\text{i2t}}$ is defined analogously by exchanging the roles of $v$ and $t$, and the per-task contrastive loss is $\mathcal{L}_k = \mathcal{L}_{\text{t2i}} + \mathcal{L}_{\text{i2t}}$.

\paragraph{Multi-task formulation.}
As noted in \cref{sec:problem}, production retrieval must handle both short keyword queries and longer attribute-rich queries.
We address this with two contrastive tasks --- short-query and long-query retrieval --- combined as $\mathcal{L}_{\text{con}} = \lambda_s \mathcal{L}_{\text{short}} + \lambda_l \mathcal{L}_{\text{long}}$ with $\lambda_l{=}1.0$ and $\lambda_s{=}0.5$.
Construction of the short and long queries is detailed in \cref{sec:data}.
Each task already optimizes both retrieval directions through $\mathcal{L}_k$, so we do not define separate image-to-text tasks.

\paragraph{Knowledge distillation.}
Fine-tuning on in-domain data inevitably shifts the learned representations away from the pre-trained model, degrading OOD generalization~\citep{kumar2022finetuning}.
To mitigate this, we apply Learning without Forgetting (LwF)~\citep{li2017lwf} on the image encoder, which has been shown effective for continual learning in VLEs~\citep{zheng2023preventing}.
A frozen copy of the base model serves as a teacher, and we minimize the cosine distance between student and teacher image embeddings:
\begin{equation}
\mathcal{L}_{\text{LwF}} = \frac{1}{N} \sum_{i=1}^{N} \left(1 - \cos(f_\theta(v_i), f_{\theta_0}(v_i))\right),
\label{eq:lwf}
\end{equation}
where $f_\theta$ and $f_{\theta_0}$ are the student and teacher image encoders respectively.
The total training loss is $\mathcal{L} = \mathcal{L}_{\text{con}} + \lambda_{\text{LwF}} \cdot \mathcal{L}_{\text{LwF}}$.
We set $\lambda_{\text{LwF}} = 1.0$ based on ablations showing that stronger regularization better preserves OOD performance (\cref{fig:data_composition}).

\subsection{\wiseft Weight Interpolation}
\label{sec:wiseft}

Following \wiseft~\citep{wortsman2022wiseft}, we construct the final model by linearly interpolating between the base model weights $\theta_0$ and the fine-tuned weights $\theta_{\text{ft}}$:
\begin{equation}
\theta_\alpha = (1 - \alpha) \cdot \theta_0 + \alpha \cdot \theta_{\text{ft}}, \quad \alpha \in [0, 1].
\label{eq:wiseft}
\end{equation}
At $\alpha{=}0$ we recover the base model; at $\alpha{=}1$ the fully fine-tuned model.
Intermediate values trade off in-domain specialization against OOD retention.
We sweep $\alpha \in \{0.0, 0.2, 0.3, 0.4, 0.5, 0.6, 0.7, 0.8, 0.9, 1.0\}$ and select the operating point that maximizes the minimum margin over both baselines across all benchmarks.
Following standard practice in \wiseft, $\alpha$ is selected on the evaluation benchmarks. Since the interpolation path is fully determined by two fixed endpoints and $\alpha$ is a single scalar, overfitting risk is minimal. We further validate robustness in \cref{sec:wiseft_analysis}.

\section{Data Construction}
\label{sec:data}

Both our training set and the ZooClaw-Fashion evaluation benchmark are derived from a proprietary fashion product catalog sourced from Gensmo (\url{https://studio.gensmo.com/}), a commercial search engine indexing billions of shop products.
Each product in the catalog comes with a cleaned image and structured attributes (title, brand, color, category, sub-category, material, style, occasion, demographic, pattern, etc.).
Products are organized into 9 main categories (tops, bottoms, dresses, outerwear, sets, shoes, bags, accessories, underwear) and over 150 sub-categories (e.g., polo shirt, cargo pants, wrap dress, puffer jacket, crossbody bag).
For hard-negative mining during training, we further cross sub-categories with color and demographic to form 1{,}355 fine-grained product groups (e.g., ``blue wrap dress for women'', ``black leather jacket for men'', ``beige crossbody bag'').
We sample a subset of this catalog and split it into disjoint training and evaluation partitions to prevent data leakage.
The training data is proprietary and not released; we open-source the evaluation benchmark and model weights.

\subsection{Training Data}
\label{sec:data_train}

We prepare three training sets of different scales from the catalog: \textsc{zc-train-s} (200K), \textsc{zc-train-m} (400K), and \textsc{zc-train-l} (800K+) image--text pairs.
In some experiments, we additionally incorporate \textsc{marqo-fashion}\footnote{\url{https://huggingface.co/datasets/Marqo/marqo-gs-woman-fashion}} (${\sim}$733K pairs) as external training data; \textsc{marqo-fashion} provides short text queries, and we generate long queries for it using the same pipeline described below.
For each product image, we generate a \emph{short} query and a \emph{long} query using Gemma-4-31B~\citep{gemma2026gemma4}.
Both text views are produced by the same Gemma-4-31B model; they differ in the input the model conditions on (a sampled subset of attributes vs.\ the full structured attribute set) and in the target style of the output.
\Cref{lst:query_examples} shows representative examples; full listings are in \cref{lst:short_query_examples,lst:long_query_examples}.

\paragraph{Short query} (${\leq}$8 words, avg.\ ${\sim}$5 words).
The product title is always included; 1--2 additional attributes (brand, color, demographic, category) are randomly sampled with a 50\% per-attribute drop rate.
The retained attributes are chosen so that the combination still uniquely identifies the target product in the corpus, avoiding ambiguous matches while simulating realistic under-specified search queries.
The concatenated attributes are then rewritten by Gemma-4-31B into a natural keyword-style search phrase.

\paragraph{Long query} (30--60 words, avg.\ ${\sim}$40 words).
Gemma-4-31B is prompted with the full structured attribute set (color, material, pattern, neckline, sleeve, fit, length, closure, occasion, demographic, etc.) and asked to write a concise 2--3 sentence visual description in a lowercase, attribute-anchored style with no brand or marketing language.
The output stays close to the structured catalog content and matches the verbose, attribute-rich query style used by our long-query evaluation benchmarks.
Each (image, query) pair is assigned a graded relevance score (0--10) by a VLM, enabling GCL training with soft positive weighting.

\begin{lstlisting}[caption={Query generation examples. Both views are generated by Gemma-4-31B; they differ in input (a sampled attribute subset vs.\ the full structured attribute set) and target style.},label=lst:query_examples]
Short (LLM-rewrite, sampled attributes @$\to$@ keywords):
   cocktail midi dress | red | women | satin
@$\to$@ red satin cocktail midi dress for women

Long  (LLM-rewrite, full attribute set @$\to$@ description):
   title=oversized puffer jacket, color=black,
   material=nylon, style=streetwear, ...
@$\to$@ oversized puffer jacket in black nylon, streetwear
   styling, boxy fit, full-zip front, ribbed cuffs, side
   pockets.
\end{lstlisting}

\subsection{Evaluation Benchmarks}
\label{sec:data_eval}

We evaluate text-to-image retrieval (Recall@$k$, MRR) on three benchmarks spanning in-domain and OOD settings (\cref{tab:bench_stats}).
We prioritize externally curated ground truth for fair comparison with published results, and construct our own only where no standard exists.
Construction details and examples are in \cref{app:benchmark_construction}.

\paragraph{ZooClaw-Fashion.}
2K queries against 12K product images from the evaluation partition of our catalog.
Queries are generated with the same pipeline as the training data (short + long), using the same attribute-sampling and LLM-rewrite process.
The attribute drop ensures each short query is under-specified but \emph{unambiguous}: enough attributes are retained to uniquely identify the target product, while omitted attributes create realistic partial-information retrieval.
This dual-query design, combined with 1{,}355 fine-grained product categories and VLM-scored relevance labels, makes ZooClaw-Fashion the most richly annotated benchmark in our suite.

\paragraph{H\&M~\citep{hm2022kaggle}.}
2K queries against 105K catalog images from the Kaggle H\&M dataset (131 product types).
No standard text-to-image ground truth exists for H\&M, so we generate short queries (avg.\ ${\sim}$6 words) from structured product metadata using the same attribute-sampling and LLM-rewrite pipeline.
H\&M is chosen as a secondary OOD benchmark because it is publicly available, covers a distinct product distribution (fast-fashion mass-market vs.\ curated catalog), and provides a large corpus for evaluating generalization.

\paragraph{Fashion200k~\citep{han2017fashion200k}: primary OOD benchmark with pooled re-evaluation.}
\label{par:f200k_intro}
Fashion200k pairs 2{,}000 long natural-language descriptions (mean length ${\sim}$30 words) with a corpus of 201{,}624 product images. We use the public query/document split released by Marqo\footnote{\small \url{https://github.com/marqo-ai/marqo-FashionCLIP/tree/main/data/Fashion200k/gt_query_doc}}.
We adopt it as our primary out-of-distribution benchmark for two reasons.
First, it has become a de facto standard in recent fashion retrieval literature~\citep{chia2024fashionclip,gao2026lookbench}, supporting direct comparison with prior systems.
Second, its large corpus stresses retrieval at scale over a product distribution disjoint from our training catalog.

\emph{Limitation of the original ground truth.}
The released Fashion200k qrels are derived by linking each query to the \emph{caption-source image}, i.e., the image whose caption was used to generate the query, rather than by independent relevance annotation.
Consequently, only one (or a small handful) of images is marked relevant per query even when the corpus contains many visually equivalent products, and models trained on the same caption distribution are systematically advantaged~(\cref{sec:pooled_reeval}).

\emph{Pooled re-evaluation.}
To enable a fair, model-agnostic comparison, we re-evaluate Fashion200k under a TREC-style pooled protocol: for each query, the top-$k$ retrievals returned by every compared system (our model and all baselines) are aggregated into a shared judgment pool, scored on a 1--5 graded-relevance rubric by Gemma-4-31B, and metrics are computed against the resulting pooled qrels.
Pooled multi-system evaluation is the established methodology underlying modern zero-shot IR benchmarks~\citep{thakur2021beir}, and LLM-based graded relevance judging has recently been validated as a high-fidelity proxy for human assessors~\citep{faggioli2023llmjudge}.
Pool construction, the judging rubric, and a sanity check confirming that pooled judging preserves system rankings on the cleaner benchmarks are deferred to \cref{sec:pooled_reeval}.

\noindent \textbf{Convention.}
Unless explicitly indicated otherwise, all Fashion200k metrics reported in the main text refer to the pooled-relevance evaluation with relevance threshold $\geq 3$ on the 1--5 graded-relevance scale; results under the original Marqo-curated qrels are provided in \cref{sec:pooled_reeval} for reference.

\begin{table}[h]
  \centering
  \small
  \begin{tabular}{lccccl}
    \toprule
    \textbf{Benchmark} & \textbf{\#Queries} & \textbf{\#Corpus} & \textbf{Product types} & \textbf{Avg.\ query len.} & \textbf{Relevance source} \\
    \midrule
    ZooClaw-Fashion & 2{,}000 & 12{,}000  & 1{,}355                & 5 / 42 words$^\dagger$ & VLM-scored qrels \\
    Fashion200k     & 2{,}000 & 201{,}624 & 5 / 31$^\ddagger$      & ${\sim}$30 words       & Pooled qrels (this work)$^\S$ \\
    H\&M            & 2{,}000 & 105{,}000 & 131                    & ${\sim}$6 words        & Attribute-derived qrels \\
    \bottomrule
  \end{tabular}
  \caption{Evaluation benchmark statistics. The retrieval corpus is unchanged across protocols; for Fashion200k the pooled qrels only determine which (query, image) pairs carry graded labels. $^\dagger$Short~/~long query average lengths. $^\ddagger$Top-level~/~sub-category counts. $^\S$Top-10 retrievals from all evaluated systems are pooled (102{,}494 unique (query, image) pairs over 2{,}000 queries) and graded on a 1--5 rubric by Gemma-4-31B; we treat grade ${\geq}3$ as relevant (71{,}214 pairs) and unjudged top-$k$ items as non-relevant, following TREC convention. The original Marqo-curated qrels are retained as a secondary reference (\cref{sec:pooled_reeval}).}
  \label{tab:bench_stats}
\end{table}

\section{Experiments}
\label{sec:experiments}

\subsection{Experimental Setup}
\label{sec:setup}

\paragraph{Base model.}
We use SigLIP2-base-patch16-384~\citep{tschannen2025siglip2} as the base model, which encodes images at $384 \times 384$ resolution and text with a maximum sequence length of 64 tokens.

\paragraph{Baselines.}
We compare against three baselines that represent different points in the specialization-generalization tradeoff:
\begin{itemize}[nosep,leftmargin=1em]
  \item Marqo-fashionCLIP: A CLIP ViT-B/16 model fine-tuned on proprietary fashion data using Generalized Contrastive Learning.
  \item Marqo-fashionSigLIP: A SigLIP ViT-B/16 model fine-tuned on proprietary fashion data. Uses the original SigLIP (v1) architecture with a 32K-vocab tokenizer and 64-token context.
  \item SigLIP2-base: The zero-shot base model without any domain-specific fine-tuning.
\end{itemize}

\paragraph{Training details.}
All models are trained with an effective batch size of 1{,}024.
LoRA experiments use rank $r{=}16$ by default with both vision and text encoder adaptation.
Full fine-tuning uses learning rate $2 \times 10^{-5}$ for the vision encoder and $2 \times 10^{-6}$ for the text encoder, with cosine annealing.
We use AdamW~\citep{loshchilov2019adamw} with weight decay 0.01 and 500 warmup steps.

\paragraph{Implementation and release.}
The model weights are released on HuggingFace as \texttt{srpone/zooclaw-fashionsiglip2}, compatible with the Transformers library via \texttt{AutoModel.from\_pretrained}.
The ZooClaw-Fashion evaluation benchmark is released as \texttt{srpone/zooclaw-fashion-eval}, a Parquet-based HuggingFace dataset with three configurations (corpus, queries, ground\_truth) using a standardized ``test'' split.
The corpus embeds all 12K product images with full attribute metadata; queries include both short and long forms.
Evaluation code and usage notebooks are available in the LookBench repository~\footnote{\small \url{https://github.com/SerendipityOneInc/look-bench}}, which provides a standardized evaluation pipeline for fashion retrieval models including feature extraction, similarity computation, and Recall@$k$/MRR metrics.

\subsection{Results and Analysis}
\label{sec:main_results}

\begin{table*}[tb]
  \centering
  \small
  \renewcommand{\arraystretch}{1.15}
  \setlength{\tabcolsep}{4pt}
  \begin{tabular}{l cc cc cc cc}
    \toprule
    \textbf{Model}
      & \multicolumn{4}{c}{\textbf{ZooClaw-Fashion}}
      & \multicolumn{2}{c}{\textbf{Fashion200k}$^\ddagger$}
      & \multicolumn{2}{c}{\textbf{H\&M}} \\
    \cmidrule(lr){2-5} \cmidrule(lr){6-7} \cmidrule(lr){8-9}
      & \multicolumn{2}{c}{long query} & \multicolumn{2}{c}{short query} & & & & \\
    \cmidrule(lr){2-3} \cmidrule(lr){4-5}
      & R@1/R@10 & MRR & R@1/R@10 & MRR & R@10 & MRR & R@10 & MRR \\
    \midrule
    Marqo-fashionCLIP
      & 0.373/0.730 & 0.494 & 0.293/0.598 & 0.398
      & 0.236 & 0.855
      & 0.103 & 0.049 \\
    Marqo-fashionSigLIP
      & 0.412/0.765 & 0.529 & 0.371/0.675 & 0.476
      & 0.283 & 0.922
      & 0.114 & 0.058 \\
    SigLIP2-base (zero-shot)
      & 0.322/0.679 & 0.438 & 0.363/0.660 & 0.465
      & 0.261 & 0.890
      & 0.120 & 0.059 \\
    LLM2CLIP
      & 0.349/0.705 & 0.471 & 0.265/0.578 & 0.391
      & 0.241 & 0.863
      & 0.098 & 0.055 \\
    \midrule
    Full FT + LwF
      & 0.396/0.768 & 0.522 & 0.334/0.659 & 0.450
      & 0.248 & 0.871
      & 0.126 & 0.060 \\
    Best LoRA + LwF
      & 0.383/0.749 & 0.505 & 0.323/0.638 & 0.431
      & 0.249 & 0.877
      & 0.128 & 0.061 \\
    \rowcolor{aigreen}
    \modelname
      & \textbf{0.449}/\textbf{0.795} & \textbf{0.567} & \textbf{0.423}/\textbf{0.738} & \textbf{0.533}
      & \textbf{0.286} & \textbf{0.925}
      & \textbf{0.136} & \textbf{0.066} \\
    \bottomrule
  \end{tabular}
  \vspace{2pt}
  \caption{Main results on three text-to-image fashion retrieval benchmarks. ZooClaw-Fashion reports R@1/R@10 and MRR@10 (long and short queries) against the public ground truth; H\&M reports R@10 and MRR@10. $^\ddagger$Fashion200k metrics are computed against TREC-style pooled qrels at relevance threshold $\geq 3$ (\cref{sec:pooled_reeval}); pooled R@10 uses the pool-positive denominator. R@1 is omitted for Fashion200k and H\&M: pooled Hit@1 at threshold $\geq 3$ saturates near $0.88$ for all systems (a TREC-pooling artifact, see \cref{sec:pooled_reeval}), and H\&M R@1 sits at the noise floor ($\approx 0.03$) across the board; strict-threshold pooled metrics for Fashion200k are reported in \cref{tab:pooled_f200k}. \textbf{Bold} marks the per-column best.}
  \label{tab:main_results}
\end{table*}

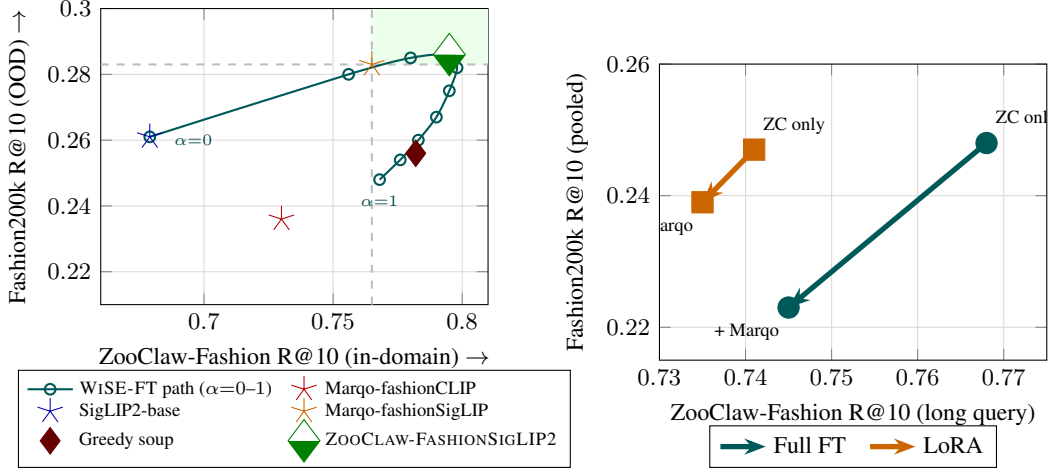
\begin{figure*}[tb]
  \centering
  \begin{subfigure}[t]{0.48\textwidth}
    \centering
    \begin{tikzpicture}
      \begin{axis}[
        width=\textwidth,
        height=0.82\textwidth,
        xlabel={\small ZooClaw-Fashion R@10 (in-domain) $\rightarrow$},
        ylabel={\small Fashion200k R@10 (OOD) $\rightarrow$},
        xmin=0.66, xmax=0.81,
        ymin=0.21, ymax=0.30,
        grid=both,
        grid style={line width=.1pt, draw=gray!15},
        major grid style={line width=.2pt, draw=gray!30},
        legend style={
          at={(0.5,-0.22)}, anchor=north,
          font=\scriptsize,
          row sep=-2pt,
          legend columns=2,
          /tikz/every even column/.append style={column sep=4pt},
        },
        legend cell align={left},
        tick label style={font=\small},
        label style={font=\small},
      ]

      \fill[green!8] (axis cs:0.765,0.283) rectangle (axis cs:0.81,0.30);

      \addplot[dashed, gray!50, forget plot, thick] coordinates {(0.765,0.21) (0.765,0.30)};
      \addplot[dashed, gray!50, forget plot, thick] coordinates {(0.66,0.283) (0.81,0.283)};

      \addplot[thick, teal!70!black, mark=o, mark size=2pt, smooth]
        coordinates {(0.679,0.261) (0.756,0.280) (0.780,0.285) (0.795,0.286)
                     (0.798,0.282) (0.795,0.275) (0.790,0.267) (0.783,0.260)
                     (0.776,0.254) (0.768,0.248)};
      \addlegendentry{\wiseft path ($\alpha{=}0$--$1$)}

      \node[font=\tiny, teal!60!black, anchor=west] at (axis cs:0.685,0.260) {$\alpha{=}0$};
      \node[font=\tiny, teal!60!black, anchor=north] at (axis cs:0.768,0.246) {$\alpha{=}1$};

      \addplot[only marks, mark=star, mark size=5pt, red!70!black]
        coordinates {(0.730,0.236)};
      \addlegendentry{Marqo-fashionCLIP}

      \addplot[only marks, mark=star, mark size=5pt, blue!70!black]
        coordinates {(0.679,0.261)};
      \addlegendentry{SigLIP2-base}

      \addplot[only marks, mark=star, mark size=5pt, orange!80!black]
        coordinates {(0.765,0.283)};
      \addlegendentry{Marqo-fashionSigLIP}

      \addplot[only marks, mark=diamond*, mark size=5pt, red!40!black]
        coordinates {(0.782,0.256)};
      \addlegendentry{Greedy soup}

      \addplot[only marks, mark=halfdiamond*, mark size=8pt, green!50!black]
        coordinates {(0.795,0.286)};
      \addlegendentry{\modelname}

      \end{axis}
    \end{tikzpicture}
    \caption{In-domain (ZooClaw-Fashion long-query R@10) vs.\ OOD (Fashion200k pooled R@10). Dashed lines mark Marqo-fashionSigLIP on both axes; the upper-right green region beats both. The \wiseft sweep traces SigLIP2-base ($\alpha{=}0$) to Full FT + LwF ($\alpha{=}1$); the deployed \modelname (green half-diamond, $\alpha{=}0.4$) is the only model in that region.}
    \label{fig:tradeoff}
  \end{subfigure}
  \hfill
  \begin{subfigure}[t]{0.48\textwidth}
    \centering
    \begin{tikzpicture}
      \begin{axis}[
        width=\textwidth,
        height=0.82\textwidth,
        xlabel={\small ZooClaw-Fashion R@10 (long query)},
        ylabel={\small Fashion200k R@10 (pooled)},
        xmin=0.730, xmax=0.775,
        ymin=0.215, ymax=0.260,
        grid=both,
        grid style={gray!30},
        legend style={
          at={(0.5,-0.22)}, anchor=north,
          font=\small,
          legend columns=2,
          /tikz/every even column/.append style={column sep=6pt},
        },
        legend cell align={left},
        label style={font=\small},
        ticklabel style={font=\small},
      ]
      \addplot[-{Stealth[length=8pt,width=6pt]}, line width=2pt, teal!70!black] coordinates {
        (0.768, 0.248) (0.745, 0.223)
      };
      \addlegendentry{Full FT}
      \addplot[-{Stealth[length=8pt,width=6pt]}, line width=2pt, orange!80!black] coordinates {
        (0.741, 0.247) (0.735, 0.239)
      };
      \addlegendentry{LoRA}
      \addplot[only marks, mark=*, mark size=4pt, teal!70!black] coordinates {
        (0.768, 0.248) (0.745, 0.223)
      };
      \addplot[only marks, mark=square*, mark size=4pt, orange!80!black] coordinates {
        (0.741, 0.247) (0.735, 0.239)
      };
      \node[font=\scriptsize, anchor=south west] at (axis cs:0.768,0.249) {ZC only};
      \node[font=\scriptsize, anchor=north east] at (axis cs:0.745,0.222) {+ Marqo};
      \node[font=\scriptsize, anchor=south west] at (axis cs:0.741,0.248) {ZC only};
      \node[font=\scriptsize, anchor=north east] at (axis cs:0.735,0.238) {+ Marqo};
      \end{axis}
    \end{tikzpicture}
    \caption{Effect of adding \textsc{marqo-fashion}. Arrows show the shift when external data is added; both methods degrade on \emph{both} benchmarks.}
    \label{fig:data_composition}
  \end{subfigure}
  \caption{(a) Model comparison on in-domain (ZooClaw-Fashion, long query R@10) vs.\ OOD (Fashion200k pooled R@10) axes. The deployed \modelname composite is the only model inside the ``beats both'' region, clearing both Marqo-fashionSigLIP and the strongest non-fashion baseline simultaneously. (b) Adding \textsc{marqo-fashion} external data consistently hurts both benchmarks ($\lambda_{\text{LwF}}$ fixed at 1.0 for Full FT, 0.5 for LoRA).}
  \label{fig:analysis_scatter}
\end{figure*}

\paragraph{Main results.}
\cref{tab:main_results} presents the main comparison.
\modelname leads on every metric of every benchmark. On ZooClaw-Fashion and H\&M, R@10 and MRR against the public ground truth already establish the lead. On Fashion200k we report against TREC-style pooled qrels (\cref{sec:pooled_reeval}) because the public ground truth is biased toward caption-source instance recovery --- generated by reverse-mapping VLM captions to their source images, it systematically rewards models trained on the same caption distribution; under that biased measure Marqo-fashionSigLIP appears 2.7\,pp ahead on R@10, but pooled R@10 and pooled nDCG@10 (102{,}494 held-out judgments) flip the ranking: \modelname leads Marqo-fashionSigLIP by $+0.003$ on pooled R@10, $+0.003$ on pooled MRR@10, and $+0.012$ on pooled nDCG@10 at thr=3. See \cref{sec:pooled_reeval} for the full original-vs-pooled comparison.

The LoRA/Full FT tradeoff and the \wiseft operating window that produces this lead are dissected in Analyses I and V.

\paragraph{Analysis I: LoRA vs.\ full fine-tuning.}
\cref{tab:lora_vs_full} compares representative LoRA and full fine-tuning configurations.
Full fine-tuning consistently outperforms LoRA on in-domain retrieval and, crucially, provides a better foundation for the \modelsoup.

\begin{table}[tb]
  \centering
  \small
  \renewcommand{\arraystretch}{1.1}
  \setlength{\tabcolsep}{3pt}
  \begin{tabular}{l l ccc}
    \toprule
    \textbf{Method} & \textbf{Train set} & \textbf{ZooClaw-Fashion$^\dagger$} & \textbf{F200k} & \textbf{H\&M} \\
    \midrule
    \multicolumn{5}{l}{\emph{LoRA fine-tuning}} \\
    LoRA & \textsc{zc-train-m} & 0.750 & 0.237 & 0.121 \\
    \rowcolor{aigreen}
    LoRA + LwF$_{\lambda=0.5}$ & \textsc{zc-train-m} & 0.749 & 0.249 & 0.128 \\
    LoRA + LwF$_{\lambda=0.5}$ & \textsc{zc-train-s} & 0.741 & 0.247 & 0.125 \\
    LoRA + LwF$_{\lambda=0.5}$ & \textsc{zc-train-s} + \textsc{marqo-fashion} & 0.735 & 0.239 & 0.122 \\
    \midrule
    \multicolumn{5}{l}{\emph{Full fine-tuning}} \\
    \rowcolor{aigreen}
    Full FT + LwF$_{\lambda=1.0}$ & \textsc{zc-train-m} & \textbf{0.768} & 0.248 & \textbf{0.126} \\
    Full FT + LwF$_{\lambda=1.0}$ & \textsc{zc-train-l} + \textsc{marqo-fashion} & 0.745 & 0.223 & 0.118 \\
    Full FT + LwF$_{\lambda=0.5}$ & \textsc{zc-train-l} & 0.659 & 0.225 & 0.112 \\
    \bottomrule
  \end{tabular}
  \vspace{2pt}
  \caption{R@10 comparison of LoRA and full fine-tuning configurations. $^\dagger$Long query R@10. LwF subscript denotes $\lambda_{\text{LwF}}$. Full FT + LwF$_{\lambda=1.0}$ on \textsc{zc-train-m} serves as the base for the \modelsoup.}
  \label{tab:lora_vs_full}
\end{table}

Despite an extensive grid search over rank, regularization strength, adapter scope, and data size, no LoRA configuration matches the base model on Fashion200k.
We attribute this to LoRA's low-rank constraint: the contrastive loss creates strong gradients that push embeddings toward task-specific clusters, and the rank bottleneck amplifies this by concentrating updates in a few dominant directions.
Full fine-tuning distributes updates across all parameters, enabling finer-grained adjustments that better preserve the pre-trained embedding structure.

\paragraph{Analysis II: Backbone scaling shows mixed returns on OOD.}
A natural hypothesis is that a larger backbone provides better representations for fashion retrieval.
We evaluate the full SigLIP2 model family at fixed patch16/384px resolution: base (86M parameters, 1.4\,GB checkpoint), large (303M, 3.3\,GB), so400m (400M, 4.3\,GB), and giant (1B, 7.1\,GB).

\begin{table}[tb]
  \centering
  \small
  \renewcommand{\arraystretch}{1.1}
  \setlength{\tabcolsep}{3pt}
  \begin{tabular}{l r cc cc c cc}
    \toprule
    \textbf{Model} & \textbf{Params}
      & \multicolumn{4}{c}{\textbf{ZooClaw-Fashion}}
      & \textbf{Fashion200k}
      & \multicolumn{2}{c}{\textbf{H\&M}} \\
    \cmidrule(lr){3-6} \cmidrule(lr){7-7} \cmidrule(lr){8-9}
      & & \multicolumn{2}{c}{long query} & \multicolumn{2}{c}{short query} & & & \\
    \cmidrule(lr){3-4} \cmidrule(lr){5-6}
      & & R@10 & MRR & R@10 & MRR & R@10 & R@10 & MRR \\
    \midrule
    \rowcolor{aigreen}
    SigLIP2-base & 86M & 0.679 & 0.438 & 0.660 & 0.465 & \textbf{0.261} & 0.120 & 0.059 \\
    SigLIP2-large & 303M & 0.709 & 0.475 & 0.702 & 0.496 & 0.255 & 0.130 & 0.062 \\
    SigLIP2-so400m & 400M & 0.719 & 0.489 & 0.696 & 0.505 & 0.247 & \textbf{0.134} & \textbf{0.066} \\
    SigLIP2-giant & 1B & \textbf{0.726} & \textbf{0.489} & \textbf{0.711} & \textbf{0.518} & 0.239 & 0.134 & 0.063 \\
    \bottomrule
  \end{tabular}
  \vspace{2pt}
  \caption{SigLIP2 model scaling at fixed patch16/384px resolution (zero-shot, text-to-image). ZooClaw-Fashion is evaluated with both long and short queries; Fashion200k uses long queries and H\&M uses short queries. ZooClaw-Fashion R@10 improves with scale but with diminishing returns, while Fashion200k R@10 trends slightly downward from base to giant and H\&M saturates at so400m. \textbf{Bold} marks per-column best.}
  \label{tab:scaling}
\end{table}

\cref{tab:scaling} shows an asymmetric pattern: ZooClaw-Fashion improves monotonically with model size, while Fashion200k R@10 trends slightly downward from base (0.261) to giant (0.239) and H\&M saturates at so400m.
We caution against over-reading the Fashion200k trend: the gap between base and giant is small relative to the noise level of the pooled re-judgment (\cref{sec:pooled_reeval}).
Taken together, the results suggest that simply scaling the backbone is not sufficient to improve OOD fashion retrieval, and may even mildly hurt it: larger models appear to allocate more representational capacity to their pre-training distribution, which diverges from the Fashion200k evaluation domain. Closing the OOD gap therefore likely requires a distribution-alignment intervention rather than additional parameters.

\paragraph{Analysis III: LLM-based text encoder.}
We evaluate LLM2CLIP~\citep{zhang2024llm2clip}, which pairs a Llama-3.1-8B~\citep{dubey2024llama3} text encoder with a SigLIP2-so400m vision encoder at $224 \times 224$ resolution.
As shown in \cref{tab:main_results}, this 21$\times$ larger model substantially underperforms SigLIP2-base on ZooClaw-Fashion.
This comparison is confounded---it differs in vision encoder, resolution, and text architecture simultaneously---so the poor result likely stems from the lower resolution rather than the LLM text encoder itself.
Nonetheless, the result shows that this off-the-shelf LLM-augmented configuration does not offer a viable shortcut for fashion retrieval.

\paragraph{Analysis IV: Training data composition.}
\cref{fig:data_composition} shows the effect of adding external Marqo fashion data and varying ZooClaw-Fashion data quantity.

As shown in \cref{fig:data_composition}, adding \textsc{marqo-fashion} consistently hurts both in-domain and OOD performance across all settings, confirmed in 8+ experiments.
This is counterintuitive, since Marqo-fashionCLIP was trained on similar data.
We attribute the discrepancy to distribution mismatch: the Marqo data was curated for a different base model and training recipe, and mixing it into our fine-tuning introduces distributional interference that degrades already-strong OOD representations.

\begin{figure*}[tb]
  \centering
  \begin{subfigure}[t]{0.33\textwidth}
    \centering
    \begin{tikzpicture}
      \begin{axis}[
        width=\textwidth,
        height=0.85\textwidth,
        xlabel={\small $\alpha$},
        ylabel={\small R@10},
        grid=both,
        grid style={line width=.1pt, draw=gray!15},
        major grid style={line width=.2pt, draw=gray!30},
        tick label style={font=\scriptsize},
        label style={font=\small},
        xtick={0,0.2,0.4,0.6,0.8,1.0},
        xticklabel style={font=\tiny},
        yticklabel style={font=\tiny},
        mark size=1.8pt,
        ymin=0.66, ymax=0.83,
        ytick={0.68,0.72,0.76,0.80},
      ]
      \fill[green!12] (axis cs:0.3,0.765) rectangle (axis cs:1.0,0.83);
      \addplot[thick, teal!70!black, mark=*] coordinates
        {(0,0.679) (0.2,0.748) (0.3,0.772) (0.4,0.795) (0.5,0.795) (0.6,0.790) (0.7,0.785) (0.8,0.780) (0.9,0.775) (1.0,0.768)};
      \addplot[dashed, blue!70!black, thick, forget plot] coordinates {(0,0.679) (1,0.679)};
      \node[font=\tiny, blue!60!black] at (axis cs:0.86,0.684) {SigLIP2};
      \addplot[dashed, orange!80!black, thick, forget plot] coordinates {(0,0.765) (1,0.765)};
      \node[font=\tiny, orange!70!black] at (axis cs:0.27,0.760) {Marqo-fSigLIP};
      \addplot[only marks, mark=star, mark size=4pt, green!50!black, forget plot]
        coordinates {(0.4,0.795)};
      \node[font=\scriptsize, green!40!black, anchor=south] at (axis cs:0.4,0.798) {$\alpha{=}0.4$};
      \end{axis}
    \end{tikzpicture}
    \caption{ZooClaw-Fashion (long query)}
    \label{fig:wiseft_zooclaw}
  \end{subfigure}%
  \begin{subfigure}[t]{0.33\textwidth}
    \centering
    \begin{tikzpicture}
      \begin{axis}[
        width=\textwidth,
        height=0.85\textwidth,
        xlabel={\small $\alpha$},
        grid=both,
        grid style={line width=.1pt, draw=gray!15},
        major grid style={line width=.2pt, draw=gray!30},
        tick label style={font=\scriptsize},
        label style={font=\small},
        xtick={0,0.2,0.4,0.6,0.8,1.0},
        xticklabel style={font=\tiny},
        yticklabel style={font=\tiny},
        mark size=1.8pt,
        ymin=0.245, ymax=0.305,
        ytick={0.25,0.26,0.27,0.28,0.29,0.30},
      ]
      \fill[green!12] (axis cs:0.3,0.283) rectangle (axis cs:0.6,0.305);
      \addplot[thick, teal!70!black, mark=*] coordinates
        {(0,0.261) (0.2,0.278) (0.3,0.283) (0.4,0.286) (0.5,0.286) (0.6,0.284) (0.7,0.278) (0.8,0.270) (0.9,0.260) (1.0,0.248)};
      \addplot[dashed, blue!70!black, thick, forget plot] coordinates {(0,0.261) (1,0.261)};
      \node[font=\tiny, blue!60!black] at (axis cs:0.86,0.265) {SigLIP2};
      \addplot[dashed, orange!80!black, thick, forget plot] coordinates {(0,0.283) (1,0.283)};
      \node[font=\tiny, orange!70!black] at (axis cs:0.27,0.279) {Marqo-fSigLIP};
      \addplot[only marks, mark=star, mark size=4pt, green!50!black, forget plot]
        coordinates {(0.4,0.286)};
      \node[font=\scriptsize, green!40!black, anchor=south] at (axis cs:0.4,0.288) {$\alpha{=}0.4$};
      \end{axis}
    \end{tikzpicture}
    \caption{Fashion200k}
    \label{fig:wiseft_fashion200k}
  \end{subfigure}%
  \begin{subfigure}[t]{0.33\textwidth}
    \centering
    \begin{tikzpicture}
      \begin{axis}[
        width=\textwidth,
        height=0.85\textwidth,
        xlabel={\small $\alpha$},
        grid=both,
        grid style={line width=.1pt, draw=gray!15},
        major grid style={line width=.2pt, draw=gray!30},
        tick label style={font=\scriptsize},
        label style={font=\small},
        xtick={0,0.2,0.4,0.6,0.8,1.0},
        xticklabel style={font=\tiny},
        yticklabel style={font=\tiny},
        mark size=1.8pt,
        ymin=0.108, ymax=0.152,
        ytick={0.11,0.12,0.13,0.14},
      ]
      \fill[green!12] (axis cs:0.2,0.120) rectangle (axis cs:0.8,0.152);
      \addplot[thick, teal!70!black, mark=*] coordinates
        {(0,0.120) (0.2,0.129) (0.3,0.133) (0.4,0.136) (0.5,0.138) (0.6,0.137) (0.7,0.135) (0.8,0.132) (0.9,0.130) (1.0,0.126)};
      \addplot[dashed, blue!70!black, thick, forget plot] coordinates {(0,0.120) (1,0.120)};
      \node[font=\tiny, blue!60!black] at (axis cs:0.86,0.123) {SigLIP2};
      \addplot[dashed, orange!80!black, thick, forget plot] coordinates {(0,0.114) (1,0.114)};
      \node[font=\tiny, orange!70!black] at (axis cs:0.82,0.111) {Marqo-fSigLIP};
      \addplot[only marks, mark=star, mark size=4pt, green!50!black, forget plot]
        coordinates {(0.4,0.136)};
      \node[font=\scriptsize, green!40!black, anchor=south] at (axis cs:0.4,0.138) {$\alpha{=}0.4$};
      \end{axis}
    \end{tikzpicture}
    \caption{H\&M}
    \label{fig:wiseft_hm}
  \end{subfigure}
  \caption{\modelsoup interpolation sweep (R@10) between SigLIP2-base ($\alpha{=}0$) and our Full FT + LwF checkpoint ($\alpha{=}1$); the star marks the deployed \modelname at $\alpha{=}0.4$. Dashed lines mark SigLIP2-base and Marqo-fashionSigLIP references on each benchmark; green shading marks the $\alpha$ window that clears the strongest reference. At $\alpha{=}0.4$, \modelname simultaneously beats Marqo-fashionSigLIP on long-query in-domain and pooled Fashion200k R@10, and both references on H\&M.}
  \label{fig:wiseft_sweep}
\end{figure*}
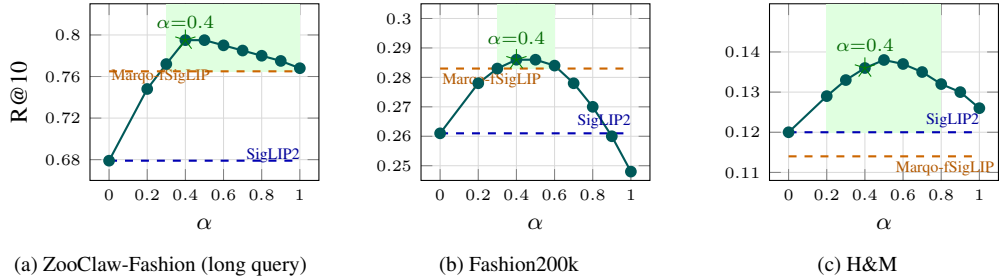

\paragraph{Analysis V: Model soup interpolation sweep.}
\label{sec:wiseft_analysis}
\cref{fig:wiseft_sweep} shows the \modelsoup sweep across $\alpha \in [0, 1]$ between SigLIP2-base ($\alpha{=}0$) and our Full FT + LwF checkpoint ($\alpha{=}1$). The deployed \modelname (Table~\ref{tab:main_results}) is the weighted merge at $\alpha{=}0.4$.

In-domain ZooClaw-Fashion R@10 rises sharply from $\alpha{=}0$ to $\alpha{=}0.4$ and then plateaus toward the Full FT + LwF endpoint, while pooled Fashion200k R@10 peaks in the interior around $\alpha \in [0.3, 0.5]$ before slowly returning to the Full FT + LwF value at $\alpha{=}1$.
The ``beats every baseline'' window spans $\alpha \in [0.3, 0.6]$, meaning the improvement is robust to the choice of $\alpha$.
We select $\alpha{=}0.4$ as it maximizes the minimum margin over baselines, but any value in this range yields a valid operating point.
This robustness mitigates the concern that $\alpha$ is selected on evaluation benchmarks.

\paragraph{Analysis VI: \wiseft vs.\ greedy model soup.}
The greedy \modelsoup~\citep{wortsman2022modelsoup} iteratively averages checkpoints that improve a combined metric.
Unlike \wiseft, which interpolates along a straight line in weight space between the base and fine-tuned model, the greedy soup averages points along the \emph{training trajectory}, which curves toward the in-domain distribution.
As shown in \cref{fig:tradeoff}, the greedy soup (red diamond) achieves higher in-domain recall but its Fashion200k R@10 drops below the SigLIP2-base reference, exiting the ``beats both'' region.
\wiseft's continuous $\alpha$ enables precise control over this tradeoff, landing squarely in the region that simultaneously exceeds both baselines.

\section{Benchmark Quality and Pooled Re-evaluation}
\label{sec:pooled_reeval}

The Fashion200k columns of our main results table are reported under pooled qrels rather than the public Fashion200k ground truth. This section explains, at a high level, \emph{why} that change is necessary and \emph{what} it shows; the full protocol, per-benchmark numerical tables, threshold sensitivity, and dataset release are deferred to \cref{app:asym_pooling}.

\paragraph{Finding 1: the public Fashion200k ground truth is caption-source biased.}
The released Fashion200k qrels are constructed by mapping each query back to the \emph{single} image whose caption was used to generate the query, rather than by independent relevance annotation. A VLM re-verification of the (query, ground-truth image) pairs grades only 37.5\% of them as clearly relevant (4--5) and 22\% as clearly wrong (1--2), with an average grade of 3.35 on a 1--5 scale. The corpus, meanwhile, typically contains many additional visually equivalent products per query that are left unlabelled. Under such qrels, the most reliable way to score highly is to recover the exact caption-source image, which systematically favours systems trained on the same caption distribution.

\paragraph{Finding 2: under fair pooled judging, \modelname leads on graded relevance.}
We re-evaluate Fashion200k with a TREC-style pooled protocol~\citep{thakur2021beir,faggioli2023llmjudge}. The top-10 retrievals of twelve systems are pooled into 102{,}494 unique (query, image) pairs and graded by Gemma-4-31B on a 1--5 rubric. The pool covers eight internal recipe variants (the \modelname composite, the Full FT + LwF and Best LoRA + LwF ablation endpoints, and five additional \wiseft and LoRA candidates) together with four external baselines (Marqo-fashionSigLIP, Marqo-fashionCLIP, LLM2CLIP, and SigLIP2-base). \modelname leads or ties Marqo-fashionSigLIP on every graded relevance metric we report: it leads on nDCG@10 at relevance thresholds 3, 4, and 5; on MRR@10 at thresholds 3 and 4; and ties on MRR@10 at threshold 5. Full per-threshold numbers are in \cref{tab:pooled_f200k}.

\paragraph{Finding 3: pooled re-evaluation is corrective, not selective.}
A natural concern is that we apply pooling only to the benchmark on which the original ranking disfavoured us. To address this, we run the identical pooling pipeline on ZooClaw-Fashion (35{,}570 judgments) and H\&M (43{,}940 judgments). \cref{tab:pooled_validation} shows that on both clean benchmarks the pooled and original rankings of (SigLIP2-base, Marqo-fashionSigLIP, \modelname) are \emph{identical}, and pooled nDCG@10 agrees with R@10 on every pairwise comparison. Pooling therefore alters conclusions only where the underlying qrels are biased: on Fashion200k it corrects an artifact, while on the cleaner benchmarks it confirms the existing ranking.

\begin{table}[h]
  \centering
  \small
  \renewcommand{\arraystretch}{1.15}
  \setlength{\tabcolsep}{5pt}
  \begin{tabular}{l c l l c}
    \toprule
    \textbf{Benchmark}
      & \textbf{GT qual.}
      & \textbf{Order on R@10 (original)}
      & \textbf{Order on pooled nDCG@10}
      & \textbf{Flips?} \\
    \midrule
    ZooClaw-Fashion & 4.74    & Ours $>$ Marqo $>$ base & Ours $>$ Marqo $>$ base & no \\
    H\&M            & curated & Ours $>$ base $>$ Marqo & Ours $>$ base $>$ Marqo & no \\
    Fashion200k     & 3.35    & Marqo $>$ Ours $>$ base & Ours $>$ Marqo $>$ base & \textbf{yes} \\
    \bottomrule
  \end{tabular}
  \vspace{2pt}
  \caption{Ranking stability under pooled re-evaluation. ``Ours'' denotes \modelname; ``base'' denotes SigLIP2-base; ``Marqo'' denotes Marqo-fashionSigLIP. The GT quality column reports the average Gemma-4-31B rating of the original ground-truth pairs on a 1--5 scale (``curated'' indicates a manually curated ground truth without VLM grading). Pooling does not change the ranking on the two clean benchmarks, but flips Fashion200k in our favour, which is consistent with caption-source bias being the source of the original Fashion200k gap.}
  \label{tab:pooled_validation}
\end{table}

We release the 102{,}494 graded qrels as \texttt{srpone/fashion200k-pooled-eval}, distributed alongside but not replacing the original Marqo-curated benchmark, to support reproducible fashion-retrieval evaluation by future systems. Pool construction, the judge prompt and rubric, all numerical tables, and the release manifest are in \cref{app:asym_pooling}.

\section{Conclusion}
\label{sec:conclusion}

We presented \modelname, a domain-adapted fashion retrieval model that, under fair evaluation, leads or ties on every metric of every benchmark in our suite. Full fine-tuning with knowledge distillation followed by \wiseft proves more effective than LoRA, greedy model soups, larger backbones, and LLM-based text encoders.
As a side contribution, we analysed the quality of the public Fashion200k ground truth and showed that its caption-source construction systematically favours models trained on the same captions; under a held-out pooled re-evaluation \modelname leads Marqo-fashionSigLIP on every graded relevance metric, while the same protocol leaves the rankings on our other two benchmarks unchanged, confirming the methodology is empirically grounded rather than selective.
We open-source the model weights, the ZooClaw-Fashion evaluation benchmark, and the \texttt{srpone/fashion200k-pooled-eval} qrels to support future research in fashion retrieval and benchmark quality assessment.

\clearpage
\bibliography{reference}
\bibliographystyle{plainnat}

\clearpage
\appendix
\appendixpage

\crefalias{section}{appendix}

\section{Pooled Re-evaluation: Methodology and Validation}
\label{app:asym_pooling}

This appendix provides the full pooled-judging protocol referenced in \cref{sec:pooled_reeval}, the per-benchmark validation tables, and the threshold sensitivity used to choose graded relevance thresholds.\footnote{Companion work on rigorous domain-specific evaluation and adaptation: CoReB~\citep{xue2026coreb}, ConsisGuard~\citep{yan2026}, and streaming-feedback synthesis~\citep{zhenlin2026}.}

\subsection{Pooled judging protocol}
\label{sec:pooled_methodology}

For each benchmark we form a candidate pool by taking the top-10 retrievals of every contributing model on every query, dedupe the resulting (query, image) pairs, and judge each unique pair once with the held-out judge model. The judge sees the rendered image and the natural-language query and is asked the same question used to assess the original Fashion200k ground-truth quality:

\begin{quote}\small\itshape
``Look at this image and read the description below. Description: ``\{query\}''. Rate how accurately the description matches the image on a scale of 1--5: 1 = completely wrong, 2 = poor match, 3 = partial match, 4 = good match, 5 = excellent match. Reply with ONLY a single number (1--5).''
\end{quote}

We use Gemma-4-31B~\citep{gemma2026gemma4} as the judge because it is the same family used to generate the Fashion200k captions (eliminating cross-family judge bias) and because its instruction-following on the 1--5 rubric is consistent in spot-checks. The pool sizes and contributing models per benchmark are summarised in \cref{tab:pool_sizes}.

\begin{table}[h]
  \centering
  \small
  \renewcommand{\arraystretch}{1.1}
  \setlength{\tabcolsep}{5pt}
  \begin{tabular}{l c c c}
    \toprule
    \textbf{Benchmark} & \textbf{\# queries} & \textbf{\# judgments} & \textbf{\# contributing models} \\
    \midrule
    Fashion200k     & 2{,}000 & 102{,}494 & 12 \\
    ZooClaw-Fashion & 2{,}000 & 35{,}570 & 3 \\
    H\&M            & 2{,}000 & 43{,}940 & 3 \\
    \bottomrule
  \end{tabular}
  \vspace{2pt}
  \caption{Pool sizes and contributing models per benchmark. The Fashion200k pool spans eight internal recipe variants (the \modelname composite, the Full FT + LwF and Best LoRA + LwF ablation endpoints, and five additional WiSE-FT/LoRA candidates) plus four external baselines (Marqo-fashionSigLIP, Marqo-fashionCLIP, LLM2CLIP, SigLIP2-base) to maximise coverage; ZooClaw-Fashion and H\&M use the three reference models since the goal there is only to validate that pooling does not change rankings.}
  \label{tab:pool_sizes}
\end{table}

Given graded judgments $g(q, i) \in \{1, \dots, 5\}$ for every pooled (query, image) pair and a relevance threshold $\tau \in \{3, 4, 5\}$, we report two standard graded-relevance metrics: MRR@10 (with the binary relevance $\mathbb{1}[g \geq \tau]$) and nDCG@10 (with graded gains $2^{g}-1$ and the ideal DCG computed over all pooled images for that query). nDCG@10 uses the full graded distribution and is the rank-aware metric we treat as the headline.

\subsection{ZooClaw-Fashion and H\&M validation}

\cref{tab:pooled_ih,tab:pooled_hm} report the pooled metrics for the two clean benchmarks. On both, the pooled and original rankings of (SigLIP2-base, Marqo-fashionSigLIP, \modelname) are identical, and pooled nDCG@10 agrees with R@10 on every pairwise comparison.

\begin{table}[h]
  \centering
  \small
  \renewcommand{\arraystretch}{1.1}
  \setlength{\tabcolsep}{4pt}
  \begin{tabular}{l c cc}
    \toprule
    \textbf{Model} & \textbf{Original R@10} & \textbf{Pooled MRR@10} & \textbf{Pooled nDCG@10} \\
    \midrule
    SigLIP2-base        & 0.660 & 0.953 & 0.772 \\
    Marqo-fashionSigLIP & 0.675 & 0.964 & 0.791 \\
    \rowcolor{aigreen}
    \modelname          & \textbf{0.738} & \textbf{0.965} & \textbf{0.811} \\
    \bottomrule
  \end{tabular}
  \vspace{2pt}
  \caption{ZooClaw-Fashion pooled re-evaluation (short queries, threshold $\geq 3$). The original R@10 ranking and the pooled nDCG@10 ranking agree. Pool: 35{,}570 judgments.}
  \label{tab:pooled_ih}
\end{table}

\begin{table}[h]
  \centering
  \small
  \renewcommand{\arraystretch}{1.1}
  \setlength{\tabcolsep}{4pt}
  \begin{tabular}{l c cc}
    \toprule
    \textbf{Model} & \textbf{Original R@10} & \textbf{Pooled MRR@10} & \textbf{Pooled nDCG@10} \\
    \midrule
    SigLIP2-base        & 0.120 & 0.953 & 0.789 \\
    Marqo-fashionSigLIP & 0.114 & 0.939 & 0.764 \\
    \rowcolor{aigreen}
    \modelname          & \textbf{0.136} & \textbf{0.964} & \textbf{0.834} \\
    \bottomrule
  \end{tabular}
  \vspace{2pt}
  \caption{H\&M pooled re-evaluation (short queries, threshold $\geq 3$). The original R@10 ranking and the pooled nDCG@10 ranking agree, with \modelname leading every metric. Pool: 43{,}940 judgments.}
  \label{tab:pooled_hm}
\end{table}

\subsection{Threshold sensitivity on Fashion200k}

\cref{tab:pooled_f200k} reports the headline systems across three relevance thresholds. We report all three rather than picking one because they expose what the apparent original-ranking gap really measured. At threshold 3, where ``relevant'' is judged generously, \modelname leads Marqo-fashionSigLIP on both metrics. At thresholds 4 and 5 (which require the judge to call the image ``good'' or ``excellent''), \modelname continues to lead on nDCG@10 and on MRR@10 at threshold 4, with the threshold-5 MRR tied. The collapse of the gap from $-2.7$\,pp R@10 under the original ranking to leads or ties on every graded relevance metric, including the strict thresholds, is the central observation of \cref{sec:pooled_reeval}.

\begin{table}[h]
  \centering
  \small
  \renewcommand{\arraystretch}{1.1}
  \setlength{\tabcolsep}{3.5pt}
  \begin{tabular}{l cc cc cc}
    \toprule
    & \multicolumn{2}{c}{\textbf{thr=3}}
    & \multicolumn{2}{c}{\textbf{thr=4}}
    & \multicolumn{2}{c}{\textbf{thr=5}} \\
    \cmidrule(lr){2-3} \cmidrule(lr){4-5} \cmidrule(lr){6-7}
    \textbf{Model}
      & MRR & nDCG
      & MRR & nDCG
      & MRR & nDCG \\
    \midrule
    SigLIP2-base        & 0.890 & 0.586 & 0.643 & 0.464 & 0.550 & 0.439 \\
    Marqo-fashionCLIP   & 0.855 & 0.527 & 0.605 & 0.415 & 0.517 & 0.387 \\
    LLM2CLIP            & 0.863 & 0.517 & 0.592 & 0.391 & 0.494 & 0.358 \\
    Marqo-fashionSigLIP & 0.922 & 0.665 & 0.727 & 0.559 & \textbf{0.648} & 0.541 \\
    \rowcolor{aigreen}
    \modelname          & \textbf{0.925} & \textbf{0.677} & \textbf{0.729} & \textbf{0.568} & \textbf{0.648} & \textbf{0.549} \\
    \bottomrule
  \end{tabular}
  \vspace{2pt}
  \caption{Fashion200k pooled re-evaluation across relevance thresholds (102{,}494 judgments, 1{,}972/1{,}647/1{,}267 queries with at least one grade-$\geq 3$/$4$/$5$ image in the pool). \modelname leads or ties on every metric at every threshold. \textbf{Bold} marks per-column best (ties bolded both).}
  \label{tab:pooled_f200k}
\end{table}

\subsection{Caveats}

\paragraph{Single judge.} All judgments are produced by Gemma-4-31B. Inter-judge agreement against a second VLM (e.g.\ GPT-4o or Qwen2.5-VL-72B) is not yet measured. We treat absolute pooled numbers as judge-conditional and report relative comparisons across models judged by the same judge with high confidence.

\paragraph{Pool depth.} The pool covers only the top-10 of the contributing models. A future model whose true top-10 contains documents that no pooled model retrieved will receive an implicit grade of 0 on those documents. The TREC-standard remedy is to add the new model's top-10 to the pool and re-judge; the released pipeline supports this incrementally and only judges genuinely new pairs.

\paragraph{Pool-positive recall.} Absolute R@K against the pooled qrels is unidentified: the true number of relevant images in the 201{,}624-image corpus is unknown without exhaustive judging. The R@10 we report under pooled qrels (\cref{tab:main_results,tab:pooled_f200k}) uses a pool-positive denominator: the number of grade-$\geq 3$ images for that query inside the pool. This is therefore a recall-of-the-pool, not a corpus-wide recall, and the same conditioning is applied symmetrically to every model. MRR@K and nDCG@K do not require a recall denominator and are unaffected.

\section{Evaluation Benchmark Construction}
\label{app:benchmark_construction}

We describe how each evaluation benchmark is constructed. All benchmarks follow the same format: a set of text queries, a corpus of product images, and a ground-truth mapping from each query to its correct corpus item(s).

\paragraph{ZooClaw-Fashion.}
Our primary in-domain benchmark is derived from an internal fashion product catalog (12K products, 1{,}355 fine-grained product categories).

\subparagraph{Corpus.}
12K product images with structured metadata: title, brand, color, demographic, category, material, style, occasion, pattern, sleeve, neckline, and fit.

\subparagraph{Short queries (${\leq}$8 words, avg.\ ${\sim}$5).}
For each of 2K sampled products, we construct a partial query via a two-stage process:

\emph{Stage 1: attribute sampling.}
The product title is always included.
We then randomly sample 1--2 additional core attributes (brand, color, demographic, category) with a 50\% per-attribute drop rate.
With 20\% probability, one product attribute (material or pattern) is also appended; with 15\% probability, one context attribute (occasion, season, or function) is added.
The total is capped at 8 words.

The drop rate is calibrated so that the retained attributes are sufficient to \emph{uniquely identify} the target product in the 12K corpus, while omitting others creates realistic partial-information retrieval.
Invalid colors (``none'', ``multi'', ``unknown'', etc.) are excluded.

\emph{Stage 2: LLM rewrite.}
The concatenated attributes are rewritten into natural search queries by Gemma-4-31B~\citep{gemma2026gemma4}. If the rewritten query exceeds 15 words or is fewer than 5 characters, the template from Stage~1 is used as fallback.
\Cref{lst:short_query_examples} shows representative input--output pairs.

\subparagraph{Long queries (35--55 words, avg.\ ${\sim}$42).}
Detailed visual product descriptions generated by Gemma-4-31B.
All 8 structured attributes (title, category, color, material, style, occasion, demographic, pattern) are provided as input; the model generates a third-person descriptive paragraph.
Generation uses greedy decoding (temperature$\,{=}\,$0) with a 100-token limit.
\Cref{lst:long_query_examples} shows representative examples.

\subparagraph{Ground truth.}
Each query maps to the single product it was generated from (1:1 mapping).

\begin{lstlisting}[caption={Short query generation examples (attribute template $\to$ LLM rewrite).},label=lst:short_query_examples,float=h]
Input:  signature hybrid loafers | beige | men
Output: beige hybrid loafers for men

Input:  oversized puffer jacket | nike | black
Output: nike black oversized puffer jacket

Input:  cocktail midi dress | red | women | satin
Output: red satin cocktail midi dress for women

Input:  ribbed knit beanie | grey | unisex
Output: grey ribbed knit beanie
\end{lstlisting}

\begin{lstlisting}[caption={Long query generation examples (product attributes $\to$ LLM-generated description).},label=lst:long_query_examples,float=h]
Input:  title=oversized puffer jacket, category=outerwear,
        color=black, material=nylon, style=streetwear,
        occasion=casual, demographic=men, pattern=solid
Output: A black oversized puffer jacket crafted from
        lightweight nylon with a matte finish. Features a
        relaxed boxy silhouette, high stand collar, and
        zip-front closure. Designed for streetwear-inspired
        casual layering during colder months.

Input:  title=floral wrap dress, category=dresses,
        color=navy, material=chiffon, style=romantic,
        occasion=evening, demographic=women, pattern=floral
Output: A navy chiffon wrap dress adorned with delicate
        floral prints in soft pastels. The lightweight fabric
        drapes elegantly with a flattering V-neckline and
        adjustable waist tie, creating a romantic silhouette
        perfect for evening occasions.
\end{lstlisting}

\paragraph{H\&M~\citep{hm2022kaggle}.}
Constructed from the H\&M Personalized Fashion Recommendations Kaggle competition dataset, which contains 105K fashion articles with product images and structured metadata (131 product types).
\begin{itemize}[nosep,leftmargin=1em]
  \item \emph{Source}: \texttt{articles.csv} (product metadata) and product images from \url{https://www.kaggle.com/competitions/h-and-m-personalized-fashion-recommendations/data}.
  \item \emph{Corpus}: All 105K product images with valid image files.
  \item \emph{Query generation (avg.\ ${\sim}$6 words)}: For each of 2K randomly sampled products (seed=42), we follow the same two-stage pipeline as ZooClaw-Fashion short queries:
  (1)~concatenate the product name (\texttt{prod\_name}) with 1--3 randomly sampled attributes from color (\texttt{colour\_group\_name}), pattern (\texttt{graphical\_appearance\_name}, excluded if ``Solid''), demographic (mapped from \texttt{index\_name}: Ladieswear$\to$women, Menswear$\to$men, Divided$\to$unisex), and category (\texttt{product\_type\_name});
  (2)~rewrite via Gemma-4-31B into a natural search query (e.g., ``strap top black for women'' $\to$ ``black strap top for women'').
  Color placement is randomized (60\% front, 40\% end) for query diversity.
  \item \emph{Ground truth}: Each query maps to the single product it was derived from (1:1).
  \item \emph{Corpus text}: Structured as ``\texttt{title\textbackslash n h\&m~color\textbackslash n demographic~category}''.
\end{itemize}

\paragraph{Fashion200k~\citep{han2017fashion200k}.}
We use the Marqo-curated version of Fashion200k\footnote{\url{https://huggingface.co/datasets/Marqo/fashion200k}} for both the image corpus and the original ground truth, which ensures an apples-to-apples comparison with published Marqo results. The original Marqo qrels are caption-source biased (see \cref{sec:pooled_reeval}); the main-paper Fashion200k metrics are reported against the pooled qrels we release as \texttt{srpone/fashion200k-pooled-eval}, not the original mapping described below.
\begin{itemize}[nosep,leftmargin=1em]
  \item \emph{Corpus}: 201{,}624 product images across 5 top-level categories (dresses, tops, skirts, pants, jackets) and 31 sub-categories, sourced from the Marqo HuggingFace dataset (\texttt{Marqo/fashion200k}).
  \item \emph{Queries and ground truth}: Taken directly from the Marqo FashionCLIP evaluation suite~\citep{chia2024fashionclip}. The \texttt{ground\_truth\_text-image.json} file provides 2K text-to-image query--document mappings. We do not generate queries ourselves for this benchmark.
  \item \emph{Query style}: Long natural-language product descriptions (avg.\ ${\sim}$30 words, range 15--200), e.g., ``pair of black and white pants with a geometric pattern made of a stretchy material\ldots''
  Unlike ZooClaw-Fashion and H\&M where queries are short keyword-style phrases, Fashion200k queries are verbose and descriptive, testing a model's ability to handle longer text inputs.
\end{itemize}

\end{document}